\documentclass[11pt, reqno]{amsart}
\usepackage{amssymb,amsthm,amsmath,epsfig,latexsym}
\usepackage{calc,times,verbatim}
\usepackage{geometry}
\usepackage{graphicx}
\usepackage{xytree}
\usepackage[T1]{fontenc}
\usepackage{extarrows}
\usepackage{color}
\usepackage{bm}
\newtheorem{Theorem}{Theorem}[section]
\newtheorem{Proposition}[Theorem]{Proposition}
\newcommand\blfootnote[1]{%
  \begingroup
  \renewcommand\thefootnote{}\footnote{#1}%
  \addtocounter{footnote}{-1}%
  \endgroup
}

\begin{document}

\title{Classified as unknown: A novel Bayesian neural network}
\author{Tianbo Yang and Tianshuo Yang}

\blfootnote{Tianbo Yang: Department of Mathematics, Haverford College, Haverford, PA 19041, USA, Email: tyang3@haverford.edu.\\
\indent Tianshuo Yang: Department of Robotics Engineering, Widener University, Chester, PA 19013, USA, Email: tyang3@widener.edu.}

\begin{abstract}
We establish estimations for the parameters of the output distribution for the softmax activation function using the probit function. As an application, we develop a new efficient Bayesian learning algorithm for fully connected neural networks, where training and predictions are performed within the Bayesian inference framework in closed-form. This approach allows sequential learning and requires no computationally expensive gradient calculation and Monte Carlo sampling. Our work generalizes the Bayesian algorithm for a single perceptron for binary classification in \cite{H} to multi-layer perceptrons for multi-class classification.
\end{abstract}
\maketitle

\vspace{-0.2in}

\section{Introduction}

Bayesian neural networks (BNNs) play an increasingly important role in machine learning. Due to their probabilistic nature, BNNs are able to measure how confident they are in their decision making, a feature that is extremely useful when dealing with noisy or confusing data such as in the case of adversarial attacks. This ability also makes them desirable in applications like medicine and autonomous driving where mistakes resulting from overconfidence can have a high cost.

In BNNs, network weights are implemented as probability distributions. These distributions are used to estimate the uncertainty in weights and predictions. However, an exact Bayesian inference for deriving the weights of an neural network is intractable due to the nonlinear nature of activation functions and the number of parameters to be estimated. Thus, the probability distributions of the weights must be approximated, typically by a Gaussion distribution. Commonly used approximation techniques include variational inference, dropout, and Kalman filters (see \cite{BCKW}, \cite{FLJHT}, \cite{GG}, \cite{G}, \cite{PF}, etc.). Since exact calculations for the parameters of the approximate distributions cannot be performed, traditional approaches to Bayesian deep learning usually employ gradient descent and Monte Carlo sampling, which make their training computationally expensive. In 2020, Marco Huber provided analytical expressions for predicting the output and for learning the weights of commonly used activation functions such as sigmoid and ReLU without the need of gradient descent and Monte Carlo sampling (see \cite{H}). However, his algorithm can only be applied to a single perceptron for binary classification due to the lack of estimations for the parameters of the output distribution of the softmax activation function.

The softmax function is an extension of the sigmoid function for more than two values. This function produces a probability distribution over multiple class labels and is typically used in the output layer of a neural network for multi-class classification. However, efficient Bayesian inference for this function is still an open problem due to its nonlinearity. Many efforts have been put in literature to find improvements to the softmax function. Brebisson and Vincent in \cite{BV} found the Taylor softmax to be a superior alternative to the regular softmax.
Liang et al. in \cite{LWLLL} proposed the soft-margin softmax that improves the discriminative property of the final layer.
Moreover, Banerjee et al. in \cite{BPGVAM} combined both functions to form the SM-Taylor softmax and showed that this new function outperforms the original Taylor softmax and soft-margin softmax for image classification tasks. All of these algorithms require gradient calculations and apply to classical neural networks, meaning that they only provide point estimates without systematically quantifying corresponding uncertainties.

In this paper, we investigate a strategy to approximate the softmax activation function. Our motivation comes from the approximation of the sigmoid function using the probit function of a single variable. By modifying certain parameters, we can use a multi-variable probit function to approximate the softmax function and vice versa. We successfully establish accurate estimations for the parameters of the output distribution of the softmax function. We then implement these estimations to a fully connected neural network to obtain analytical expressions for predicting the output and for learning the weights.  This novel algorithm is efficient and allows sequential learning because its training and prediction are performed within the Bayesian inference framework without the need of gradient descent and Monte Carlo sampling. It generalizes the Bayesian algorithm for a single perceptron for binary classification in \cite{H} to multi-layer perceptrons for multi-class classification.

The paper is structured as follows: In the next section, we demonstrate analytical estimations for the softmax function. In Section 3, we define the forward pass of the proposed Bayesian algorithm for estimating the output distributions in fully connected neural networks. We then provide the backward pass procedure for updating the weights in Section 4. In Section 5, we apply this novel approach to synthetic data and analyze the experiment results. Section 6 concludes the paper and shares our plan for future work. Section 7 is an appendix containing additional proofs.

\section{Softmax function}

The softmax function is widely used in artificial neural networks for multi-class classification. However, its computational complexity is often questioned in literature. In this section, we investigate the approximation of this activation function using the probit function.

Let $\underline{z}=[z_1 \,\, \ldots \,\, z_n]\sim N\left(\mu_{\underline{z}}, \Sigma_{\underline{z}}\right)$ be $n$ Gaussian random variables with mean vector $\mu_{\underline{z}}$ and covariance matrix $\Sigma_{\underline{z}}$, where $\left(\mu_{\underline{z}}\right)_j=\mu_{z_j}$ and
$\left(\Sigma_{\underline{z}}\right)_{jj}=\sigma_{z_j}^2, \left(\Sigma_{\underline{z}}\right)_{ij}=0$ for $1\leq i, j\leq n$ and $i\neq j$.

Define
$$
\underline{y}=\left[y_1 \,\, \cdots \,\, y_{n-1}\right]={\rm softmax}(\underline{z})=\left[\begin{array}{ccc}\frac{{\rm exp}(z_1)}{\sum_{1\leq \tau\leq n} {\rm exp}(z_{\tau})}&\cdots& \frac{{\rm exp}(z_{n-1})}{\sum_{1\leq \tau\leq n} {\rm exp}(z_{\tau})}\end{array}\right].
$$
Let $y_n=1-\sum_{j=1}^{n-1}y_j$. Then we have a probability distribution over the $n$ class labels. We use the arg max function to generate the class label, where
$$
{\rm arg\,max}(\underline{y})=j \quad {\rm where} \quad y_{j}={\rm max}\{y_{\tau} \,|\, 1\leq \tau\leq n\}.
$$
One would like to estimate the mean vector $\mu_{\underline{y}}$ and the covariance matrices $\Sigma_{\underline{y}}$ and $\Sigma_{\underline{z}\underline{y}}$ in terms of $\mu_{\underline{z}}$ and $\Sigma_{\underline{z}}$.

For each $j$ with $1\leq j\leq n-1$, write $y_j=s\left(\underline{z}, j\right)=\frac{{\rm exp}(z_j)}{\sum_{1\leq \tau\leq n} {\rm exp}(z_{\tau})}$. Then for $1\leq i\leq n$, we have
\begin{eqnarray*}
\mu_{y_j}&=&E\{y_j\}=E\{s\left(\underline{z}, j\right)\},\\
\sigma_{y_iy_j}&=&E\{(y_i-\mu_{y_i})(y_j-\mu_{y_j})\}=E\{y_iy_j\}-\mu_{y_i}\mu_{y_j}=E\{s\left(\underline{z}, i\right)s\left(\underline{z}, j\right)\}-\mu_{y_i}\mu_{y_j},\\
\sigma_{z_iy_j}&=&E\{\left(z_i-\mu_{z_i})(y_j-\mu_{y_j}\right)\}=E\{z_iy_j\}-\mu_{z_i}\mu_{y_j}=E\{z_i s\left(\underline{z}, j\right)\}-\mu_{z_i}\mu_{y_j}.
\end{eqnarray*}

To estimate the above parameters, we define the probit function of $n-1$ variables:
\begin{equation}\label{DefinitionPhi}
\Phi\left(\underline{x}; \mu_{\underline{x}}, \Sigma_{\underline{x}}\right)=\int_{-\infty}^{x_{n-1}}\cdots\int_{-\infty}^{x_1}
N\left(\underline{t}; \mu_{\underline{x}}, \Sigma_{\underline{x}}\right)dt_1\cdots dt_{n-1},
\end{equation}
where $\underline{x}=[x_1 \,\, \ldots \,\, x_{n-1}]\sim N\left(\mu_{\underline{x}}, \Sigma_{\underline{x}}\right)$ are
Gaussian random variables with mean vector $\mu_{\underline{x}}$ and covariance matrix $\Sigma_{\underline{x}}$. Let
\begin{equation}\label{mean}
\mu_{\underline{0}}=\left[\begin{array}{ccc}0&\cdots&0\end{array}\right]_{1\times (n-1)}
\end{equation}
and
\begin{equation}\label{variance}
\Sigma_{\underline{0}}=\left[\begin{array}{cccc}1&\rho_{12}&\ldots&\rho_{1(n-1)}\\ \vdots&\vdots&\vdots&\vdots\\ \rho_{1(n-1)}&\rho_{2(n-1)}&\ldots&1\end{array}\right]_{(n-1)\times (n-1)}.
\end{equation}
One can modify parameters $\lambda$ and $\rho_{ij}$, where $1\leq i<j\leq n-1$, so that $\Phi\left(\lambda\theta_{\tau}, 1\leq \tau\leq n, \tau\neq j; \mu_{\underline{0}}, \Sigma_{\underline{0}}\right)$ can be used to approximate
$$
s\left(\theta_{\tau}, 1\leq \tau\leq n, \tau\neq j\right)=\frac{1}{1+\sum_{1\leq \tau\leq n, \tau\neq j}{\rm exp}\left(-\theta_{\tau}\right)}.
$$
Therefore,
$$
y_j=s\left(\underline{z}, j\right)=s\left(z_j-z_{\tau}, 1\leq \tau\leq n, \tau\neq j\right)\approx \Phi\left(\lambda\left(z_j-z_{\tau}\right), 1\leq i\leq n, {\tau}\neq j; \mu_{\underline{0}}, \Sigma_{\underline{0}}\right).
$$

Now we have the following estimate for $\mu_{y_j}$:
\begin{eqnarray*}
&&\mu_{y_j}=E\{s\left(\underline{z}, j\right)\}
=\int_{\mathbb{R}}\cdots\int_{\mathbb{R}}s\left(\underline{z}, j\right)
N\left(\underline{z}; \mu_{\underline{z}}, \Sigma_{\underline{z}}\right)dz_1\cdots dz_n\\
&\approx&\int_{\mathbb{R}}\cdots\int_{\mathbb{R}}\Phi\left(\lambda\left(z_j-z_{\tau}\right), 1\leq \tau\leq n, \tau\neq j; \mu_{\underline{0}}, \Sigma_{\underline{0}}\right)N\left(\underline{z}; \mu_{\underline{z}}, \Sigma_{\underline{z}}\right)dz_1\cdots dz_n\\
&=&\int_{\mathbb{R}}\cdots\int_{\mathbb{R}}\Phi\left(\widehat{z}_j-\widehat{z}_{\tau}, 1\leq \tau\leq n, \tau\neq j; \mu_{\underline{0}}, \Sigma_{\underline{0}}\right)N\left(\widehat{\underline{z}}; \lambda\mu_{\underline{z}}, \lambda^2\Sigma_{\underline{z}}\right)d\widehat{z}_1\cdots d\widehat{z}_n\\
&=&\Phi\left(\frac{\lambda\left(\mu_{z_j}-\mu_{z_{\tau}}\right)}{\sqrt{1+\lambda^2\sigma_{z_j}^2+\lambda^2\sigma_{z_{\tau}}^2}}, 1\leq \tau\leq n, \tau\neq j; \mu_{\underline{0}}, \Sigma_{\underline{0}}'\right),
\end{eqnarray*}
where $\widehat{z}_{\tau}=\lambda z_{\tau}$ for $1\leq \tau\leq n$ and the last equality follows from Proposition \ref{phi} (1) in Section~7 Appendix.
\smallskip

Observe
\begin{eqnarray*}
&&E\left\{z_jy_j\right\}=E\left\{z_j\,s\left(\underline{z}, j\right)\right\}=
\int_{\mathbb{R}}\cdots\int_{\mathbb{R}}z_j\,s\left(\underline{z}, j\right)
N\left(\underline{z}; \mu_{\underline{z}}, \Sigma_{\underline{z}}\right)dz_1\cdots dz_n\\
&=&\int_{\mathbb{R}}\cdots\int_{\mathbb{R}}\left(\sigma_{z_j}^2\left(\frac{z_j-\mu_{z_j}}{\sigma_{z_j}^2}\right)+\mu_{z_j}\right)
\,s\left(\underline{z}, j\right)N\left(\underline{z}; \mu_{\underline{z}}, \Sigma_{\underline{z}}\right)dz_1\cdots dz_n\\
&=&\sigma_{z_j}^2\int_{\mathbb{R}}\cdots\int_{\mathbb{R}}\left(\frac{z_j-\mu_{j}}{\sigma_{z_j}^2}\right)
\,s\left(\underline{z}, j\right)N\left(\underline{z}; \mu_{\underline{z}}, \Sigma_{\underline{z}}\right)dz_1\cdots dz_n +\mu_{z_j}\mu_{y_j}\\
&=&\sigma_{z_j}^2\int_{\mathbb{R}}\cdots\int_{\mathbb{R}}\left(\frac{\partial \,s\left(\underline{z}, j\right)}{\partial z_j}\right)
N\left(\underline{z}; \mu_{\underline{z}}, \Sigma_{\underline{z}}\right)dz_1\cdots dz_n +\mu_{z_j}\mu_{y_j}\\
&=&\sigma_{z_j}^2\int_{\mathbb{R}}\cdots\int_{\mathbb{R}}\left(\sum_{1\leq i\leq n, i\neq j}
\frac{\partial \,s\left(\underline{z}, j\right)}{\partial (z_j-z_i)}\right)
N\left(\underline{z}; \mu_{\underline{z}}, \Sigma_{\underline{z}}\right)dz_1\cdots dz_n +\mu_{z_j}\mu_{y_j}\\
&=&\sigma_{z_j}^2\sum_{1\leq i\leq n, i\neq j}E\left\{\frac{\partial \,s\left(\underline{z}, j\right)}{\partial (z_j-z_i)}\right\}
+\mu_{z_j}\mu_{y_j}.
\end{eqnarray*}
For $i\neq j$,
\begin{eqnarray*}
&&E\left\{z_iy_j\right\}=E\left\{z_i\,s\left(\underline{z}, j\right)\right\}=
\int_{\mathbb{R}}\cdots\int_{\mathbb{R}}z_i\,s\left(\underline{z}, j\right)
N\left(\underline{z}; \mu_{\underline{z}}, \Sigma_{\underline{z}}\right)dz_1\cdots dz_n\\
&=&\int_{\mathbb{R}}\cdots\int_{\mathbb{R}}\left(\sigma_{z_i}^2\left(\frac{z_i-\mu_{z_i}}{\sigma_{z_i}^2}\right)+\mu_{z_i}\right)
\,s\left(\underline{z}, j\right)N\left(\underline{z}; \mu_{\underline{z}}, \Sigma_{\underline{z}}\right)dz_1\cdots dz_n\\
&=&\sigma_{z_i}^2\int_{\mathbb{R}}\cdots\int_{\mathbb{R}}\left(\frac{z_i-\mu_{i}}{\sigma_{z_i}^2}\right)
\,s\left(\underline{z}, j\right)N\left(\underline{z}; \mu_{\underline{z}}, \Sigma_{\underline{z}}\right)dz_1\cdots dz_n +\mu_{z_i}\mu_{y_j}
\end{eqnarray*}
\begin{eqnarray*}
&=&\sigma_{z_i}^2\int_{\mathbb{R}}\cdots\int_{\mathbb{R}}\left(\frac{\partial \,s\left(\underline{z}, j\right)}{\partial z_i}\right)
N\left(\underline{z}; \mu_{\underline{z}}, \Sigma_{\underline{z}}\right)dz_1\cdots dz_n +\mu_{z_i}\mu_{y_j}\\
&=&-\sigma_{z_i}^2\int_{\mathbb{R}}\cdots\int_{\mathbb{R}}\left(\frac{\partial \,s\left(\underline{z}, j\right)}{\partial \left(z_j-z_i\right)}\right)N\left(\underline{z}; \mu_{\underline{z}}, \Sigma_{\underline{z}}\right)dz_1\cdots dz_n+\mu_{z_i}\mu_{y_j}\\
&=&-\sigma_{z_i}^2E\left\{\frac{\partial \,s\left(\underline{z}, j\right)}{\partial \left(z_j-z_i\right)}\right\}
+\mu_{z_i}\mu_{y_j}.
\end{eqnarray*}

Since
$$
y_j(1-y_j)=\sum_{1\leq i\leq n, i\neq j}\frac{\partial \,s\left(\underline{z}, j\right)}{\partial (z_j-z_i)} \,\,\, {\rm and} \,\,\,
y_iy_j=\frac{\partial \,s\left(\underline{z}, j\right)}{\partial (z_j-z_i)} \,\, {\rm for} \,\, i\neq j,
$$
we have
\begin{eqnarray*}
\sigma_{y_j}^2&=&E\left\{y_j^2\right\}-\mu_{y_j}^2
=E\left\{y_j-y_j\left(1-y_j\right)\right\}-\mu_{y_j}^2\\
&=&\mu_{y_j}-\mu_{y_j}^2-\sum_{1\leq i\leq n, i\neq j}E\left\{\frac{\partial \,s\left(\underline{z}, j\right)}{\partial (z_j-z_i)}\right\},\\
\sigma_{y_iy_j}&=&E\left\{y_iy_j\right\}-\mu_{y_i}\mu_{y_j}\\
&=&-\mu_{y_i}\mu_{y_j}+E\left\{\frac{\partial \,s\left(\underline{z}, j\right)}{\partial (z_j-z_i)}\right\}, \quad i\neq j, \\
\sigma_{z_jy_j}&=&E\{z_jy_j\}-\mu_{z_j}\mu_{y_j}=\sigma_{z_j}^2\sum_{1\leq i\leq n, i\neq j}E\left\{\frac{\partial \,s\left(\underline{z}, j\right)}{\partial (z_j-z_i)}\right\},\\
\sigma_{z_iy_j}&=&E\left\{z_iy_j\right\}-\mu_{z_i}\mu_{y_j}=-\sigma_{z_i}^2E\left\{\frac{\partial \,s\left(\underline{z}, j\right)}{\partial \left(z_j-z_i\right)}\right\}, \quad i\neq j.
\end{eqnarray*}

We are left to estimate $E\left\{\frac{\partial \,s\left(\underline{z}, j\right)}{\partial \left(z_j-z_i\right)}\right\}$ for $1\leq i\leq n$ and $i\neq j$:
\begin{eqnarray*}
&&E\left\{\frac{\partial \,s\left(\underline{z}, j\right)}{\partial \left(z_j-z_i\right)}\right\}
=\int_{\mathbb{R}}\cdots\int_{\mathbb{R}}\frac{\partial \,s\left(\underline{z}, j\right)}{\partial (z_j-z_i)}
N\left(\underline{z}; \mu_{\underline{z}}, \Sigma_{\underline{z}}\right)dz_1\cdots dz_n\\
&\approx&\int_{\mathbb{R}}\cdots\int_{\mathbb{R}}\frac{\partial \,\Phi\left(\lambda\left(z_j-z_{\tau}\right), 1\leq \tau\leq n, \tau\neq j; \mu_{\underline{0}}, \Sigma_{\underline{0}}\right)}{\partial \left(z_j-z_i\right)}N\left(\underline{z}; \mu_{\underline{z}}, \Sigma_{\underline{z}}\right)dz_1\cdots dz_n\\
&=&\lambda\int_{\mathbb{R}}\cdots\int_{\mathbb{R}}\frac{\partial \,\Phi\left(\lambda\left(z_j-z_{\tau}\right), 1\leq \tau\leq n, \tau\neq j; \mu_{\underline{0}}, \Sigma_{\underline{0}}\right)}{\partial \lambda\left(z_j-z_i\right)}N\left(\underline{z}; \mu_{\underline{z}}, \Sigma_{\underline{z}}\right)dz_1\cdots dz_n\\
&=&\lambda\int_{\mathbb{R}}\cdots\int_{\mathbb{R}}\frac{\partial \,\Phi\left(\widehat{z}_j-\widehat{z}_{\tau}, 1\leq \tau\leq n, \tau\neq j; \mu_{\underline{0}}, \Sigma_{\underline{0}}\right)}{\partial \left(\widehat{z}_j-\widehat{z}_i\right)}N\left(\widehat{\underline{z}}; \lambda\mu_{\underline{z}}, \lambda^2\Sigma_{\underline{z}}\right)d\widehat{z}_1\cdots d\widehat{z}_n\\
&=&\frac{1}{\sqrt{1+\lambda^2\sigma_{z_j}^2+\lambda^2\sigma_{z_i}^2}}
\frac{\partial\Phi}{\partial\left(z_j-z_i\right)}\left(\frac{\lambda\left(\mu_{z_j}-\mu_{z_{\tau}}\right)}
{\sqrt{1+\lambda^2\sigma_{z_j}^2+\lambda^2\sigma_{z_{\tau}}^2}}, 1\leq \tau\leq n, \tau\neq j; \mu_{\underline{0}}, \Sigma_{\underline{0}}'\right),
\end{eqnarray*}
where $\widehat{z}_{\tau}=\lambda z_{\tau}$ for $1\leq \tau\leq n$ and the last equality follows from Proposition \ref{phi} (2) in Section~7 Appendix.

When $n=2$, the softmax function of binary class can be reduced to the sigmoid function by
$a=z_1-z_2$ and $y=y_1=\frac{e^{z_1}}{e^{z_1+z_2}}=\frac{1}{1+e^{-a}}=s\left(a\right)$. Then $\mu_{a}=\mu_{z_1}-\mu_{z_2}$ and
$\sigma_{a}^2=\sigma_{z_1}^2+\sigma_{z_2}^2$. By our formulas:
\begin{eqnarray*}
\mu_{y}&\approx& \Phi\left(\frac{\lambda\left(\mu_{z_1}-\mu_{z_{2}}\right)}{\sqrt{1+\lambda^2\sigma_{z_1}^2+\lambda^2\sigma_{z_{2}}^2}}; 0, 1\right)=\Phi\left(\frac{\lambda\mu_{a}}{\sqrt{1+\lambda^2\sigma_{a}^2}}; 0, 1\right),\\
\sigma^2_y&\approx& \mu_{y_1}-\mu_{y_1}^2-\frac{1}{\sqrt{1+\lambda^2\sigma_{z_1}^2+\lambda^2\sigma_{z_2}^2}}
\frac{d\Phi}{d\left(z_1-z_2\right)}\left(\frac{\lambda\left(\mu_{z_1}-\mu_{z_{2}}\right)}
{\sqrt{1+\lambda^2\sigma_{z_1}^2+\lambda^2\sigma_{z_{2}}^2}}; 0, 1\right)\\
&=& \mu_{y}-\mu_{y}^2-\frac{1}{\sqrt{1+\lambda^2\sigma_{a}^2}}
\frac{d\Phi}{da}\left(\frac{\lambda\mu_{a}}
{\sqrt{1+\lambda^2\sigma_{a}^2}}; 0, 1\right),\\
\sigma_{a y}&=&\sigma_{z_1 y_1}-\sigma_{z_2 y_1}=\frac{\sigma_{z_1}^2+\sigma_{z_2}^2}{\sqrt{1+\lambda^2\sigma_{z_1}^2+\lambda^2\sigma_{z_2}^2}}
\frac{d\Phi}{d\left(z_1-z_2\right)}\left(\frac{\lambda\left(\mu_{z_1}-\mu_{z_{2}}\right)}
{\sqrt{1+\lambda^2\sigma_{z_1}^2+\lambda^2\sigma_{z_{2}}^2}}; 0, 1\right)\\
&=&\frac{\sigma_{a}^2}{\sqrt{1+\lambda^2\sigma_{a}^2}}\frac{d\Phi}{da}\left(\frac{\lambda\mu_{a}}
{\sqrt{1+\lambda^2\sigma_{a}^2}}; 0, 1\right),
\end{eqnarray*}
which coincide with the estimations given by Marco Huber in \cite{H} for the sigmoid function.
\smallskip

\section{Bayesian Algorithm: Forward Pass}
\medskip

Let $D=\left\{\underline{x}_k, \underline{y}_k\right\}_{k=1}^m$ be a training dataset consisting of $m$ identically distributed independent training instances $(\underline{x}_k, \underline{y}_k)$ with input/features $\underline{x}_k=[x_{k1} \, \ldots \, x_{kd}]\in \mathbb{R}^d$ and output $\underline{y}_k\in \mathbb{R}^{N-1}$ in the case of $N$ class classification. The output $\underline{y}_k=[0\,\cdots\, 0\, 1\, 0\, \cdots\, 0]_{1\times (N-1)}$, where the $j$th entry is 1 if the data represents the $j$th class for $1\leq j\leq N-1$ (other entries are just zero). If the data represents the $N$th class, then $\underline{y}_k=[0\,\cdots\, 0]_{1\times (N-1)}$ has only zero entries.

The diagram of a many-layered network has the following form:
$$
\xrightarrow{\underline{x}=\underline{y}^0} \underset{\rm layer \, 1}{\underbrace{\fbox{$\begin{array}{c} \underline{w}^1 \\ \underline{w}_0^1\end{array}$} \xrightarrow{\underline{z}^1}\fbox{$f^1$}}}
\xrightarrow{\underline{y}^1} \underset{\rm layer \, 2}{\underbrace{\fbox{$\begin{array}{c} \underline{w}^2 \\ \underline{w}_0^2\end{array}$} \xrightarrow{\underline{z}^2}\fbox{$f^2$}}}\xrightarrow{\underline{y}^2} \ldots \xrightarrow{\underline{y}^{L-1}}
\underset{\rm layer \, L}{\underbrace{\fbox{$\begin{array}{c} \underline{w}^L \\ \underline{w}_0^L\end{array}$} \xrightarrow{\underline{z}^L}\fbox{$f^L$}}}\xrightarrow{\underline{y}^L}
\fbox{Loss}
$$
For $1\leq\ell\leq L$, each layer $\ell$ has input $\underline{y}^{\ell-1}\in \mathbb{R}^{n^{\ell-1}}$ and $\underline{z}^{\ell}={\underline{y}^{\ell-1}}{\underline{w}^{\ell}}+\underline{w}_0^{\ell}\in \mathbb{R}^{n^{\ell}}$,  where $\underline{w}^{\ell}\in \mathbb{R}^{n^{\ell-1}\times n^\ell}$ comprises the weights and $\underline{w}_0^{\ell}\in \mathbb{R}^{1\times n^\ell}$ consists of the biases. We may assume $\underline{w}_0^{\ell}=\underline{0}$ by including the bias vector into the weight matrix: $\underline{z}^{\ell}=\widehat{\underline{y}}^{\ell-1}\,\widehat{\underline{w}}^{\ell}$ with $\widehat{\underline{y}}^{\ell-1}=\left[\underline{y}^{\ell-1} \,\ 1\right]$ and $\widehat{\underline{w}}^{\ell}=\left[\begin{array}{c}\underline{w}^{\ell}\\\underline{w}_0^{\ell}\end{array}\right]$.
The output $\underline{y}^{\ell}=f^{\ell}\left(\underline{z}^{\ell}\right)\in \mathbb{R}^{n^{\ell}}$ for $1\leq \ell\leq L-1$ with the nonlinear activation function $f^{\ell}$ applied element-wise to the entries of $\underline{z}^{\ell}$. Let $N=n^L$. Then the last output $\underline{y}^{L}=[y_1^L\,\cdots\,y_{N-1}^L]\in \mathbb{R}^{N-1}$ with $y_j^L=s\left(\underline{z}^L, j\right)=\frac{{\rm exp}\left(z_j^L\right)}{\sum_{1\leq \tau\leq n} {\rm exp}\left(z_{\tau}^L\right)}$ for $1\leq j\leq N-1$.

Write $\underline{w}^{\ell}=\left[\begin{array}{ccc}\underline{w}_1^{\ell}&\cdots&\underline{w}_{n^{\ell}}^{\ell}\end{array}\right]$ and
$\underline{z}^{\ell}=\left[\begin{array}{ccc}z_1^{\ell}&\cdots&z_{n^{\ell}}^{\ell}\end{array}\right]$ for $1\leq \ell\leq L$.
Assume $\underline{w}_1^{\ell}, \ldots, \underline{w}_{n^{\ell}}^{\ell}$ are independent with $\underline{w}_j^{\ell}\sim N\left(\mu^{\underline{w}_j^{\ell}}, \Sigma^{\underline{w}_j^{\ell}}\right)$ for $1\leq j\leq n^{\ell}$,
where $\mu^{\underline{w}_j^{\ell}}$ is the mean vector and $\Sigma^{\underline{w}_j^{\ell}}$ is the covariance matrix.
Since $\underline{z}^{\ell}=\underline{y}^{\ell-1}\underline{w}^{\ell}$, we have that $\underline{z}^{\ell}\sim N\left(\mu_{\underline{z}^{\ell}}, \Sigma_{\underline{z}^{\ell}}\right)$ with mean and variance:
\begin{eqnarray}\label{zz}
\left(\mu_{\underline{z}^{\ell}}\right)_j&=&\mu_{z_j^{\ell}}=\underline{y}^{\ell-1}\mu^{\underline{w}_j^{\ell}}\nonumber\\
\left(\Sigma_{\underline{z}^{\ell}}\right)_{jj}
&=&\sigma_{z_j^{\ell}}^2=\underline{y}^{\ell-1}\Sigma^{\underline{w}_j^{\ell}}\left(\underline{y}^{\ell-1}\right)^T, \,
\left(\Sigma_{\underline{z}^{\ell}}\right)_{ij}=0 \,\, {\rm for} \,\, 1\leq i, j\leq n^{\ell} \,\, {\rm and} \,\, i\neq j.
\end{eqnarray}
\newpage

Due to the nonlinearity introduced by the activity function $f^{\ell}$, an exact calculation of the probability density function is only possible in few cases. We apply the usual Bayesian neural network assumption that the probability density function can be approximated well by a Gaussian distribution, i.e., $\underline{y}^{\ell}=f^{\ell}(\underline{z}^{\ell})\approx N\left(\mu_{\underline{y}^{\ell}}, \Sigma_{\underline{y}^{\ell}}\right)$ with the mean vector $\mu_{\underline{y}^{\ell}}$ and the covariance matrix $\Sigma_{\underline{y}^{\ell}}$.
\medskip

Recall a piece-wise linear activation function is defined as $f(z)={\rm max}(\alpha z, \beta z)$ with $\alpha\in[0, 1]$ and $0\leq \alpha\leq \beta$. When $\alpha=0$ and $\beta=1$, this definition reduces to the rectified linear unit (ReLU) as a special case. In general, we use piece-wise linear functions in internal (``hidden") layers and the softmax function for the output layer for $N$ class classification. By Section 2 above and the work for piece-wise linear functions in \cite{H}, we have the following estimations for each layer $1\leq \ell\leq L$:

\begin{Proposition}
\begin{enumerate}
\item \, Let $1\leq \ell\leq L-1$. Then $\underline{y}^{\ell}=\left[y_1^{\ell}\,\,\cdots\,\,y_{n^{\ell}}^{\ell}\right]\in \mathbb{R}^{n^{\ell}}$ with $y_j^{\ell}={\rm max}\left\{\alpha z_j^{\ell}, \beta z_j^{\ell}\right\}$ for $1\leq j\leq n^{\ell}$. Set $E_1=\mu_{z_j^{\ell}}$ and $E_2=\mu_{z_j^{\ell}}^2+\sigma_{z_j^{\ell}}^2$, one has the following estimations for $1\leq j\leq n^{\ell}$:
\begin{eqnarray*}
\left(\mu_{\underline{y}^{\ell}}\right)_j&=&\mu_{y_j^{\ell}}=\alpha E_1 +\left(\beta-\alpha\right)\left(E_1 \Phi\left(\frac{\mu_{z_j^{\ell}}}{\sigma_{z_j^{\ell}}}; 0, 1\right)+
\sigma_{z_j^{\ell}}N\left(\frac{\mu_{z_j^{\ell}}}{\sigma_{z_j^{\ell}}}; 0, 1\right)\right),\\
\left(\Sigma_{\underline{y}^{\ell}}\right)_{jj}&=&\sigma_{y_j^{\ell}}^2=\alpha^2E_2+\left(\beta^2-\alpha^2\right)\left(E_2 \Phi\left(\frac{\mu_{z_j^{\ell}}}{\sigma_{z_j^{\ell}}}; 0, 1\right)+\mu_{z_j^{\ell}}\sigma_{z_j^{\ell}}N\left(\frac{\mu_{z_j^{\ell}}}{\sigma_{z_j^{\ell}}}; 0, 1\right)\right)-\mu_{y_j^{\ell}}^2,\\
\left(\Sigma_{\underline{y}^{\ell}}\right)_{ij}&=&\sigma_{y_i^{\ell}y_j^{\ell}}=0, \quad i\neq j,\\
\left(\Sigma_{\underline{z}^{\ell}\underline{y}^{\ell}}\right)_{jj}&=&\sigma_{z_j^{\ell}y_j^{\ell}}=\alpha E_2+\left(\beta-\alpha\right)\left(E_2 \Phi\left(\frac{\mu_{z_j^{\ell}}}{\sigma_{z_j^{\ell}}}; 0, 1\right)+\mu_{z_j^{\ell}}\sigma_{z_j^{\ell}}N\left(\frac{\mu_{z_j^{\ell}}}{\sigma_{z_j^{\ell}}}; 0, 1\right)\right)
-\mu_{z_j^{\ell}}\mu_{y_j^{\ell}},\\
\left(\Sigma_{\underline{z}\underline{y}^{\ell}}\right)_{ij}&=&\sigma_{z_i^{\ell}y_j^{\ell}}=0, \quad i\neq j.
\end{eqnarray*}
\item \, Let $\ell=L$ and $N=n^L$. Then $\underline{y}^{L}=\left[y_1^{L}\,\,\cdots\,\,y_{N-1}^{L}\right]\in \mathbb{R}^{N-1}$ with $y_j^L=s\left(\underline{z}^L, j\right)=\frac{{\rm exp}\left(z_j^L\right)}{\sum_{1\leq \tau\leq N} {\rm exp}\left(z_{\tau}^L\right)}$
    for $1\leq j\leq N-1$. One the following estimations for $1\leq j\leq N-1$:
\begin{eqnarray*}
\left(\mu_{\underline{y}^{L}}\right)_j&=&\mu_{y_j^L}\approx\Phi\left(\frac{\lambda\left(\mu_{z_j^L}-\mu_{z_{\tau}^L}\right)}{\sqrt{1+\lambda^2\sigma_{z_j^L}^2+\lambda^2\sigma_{z_{\tau}^L}^2}}, 1\leq \tau\leq N, \tau\neq j; \mu_{\underline{0}}, \Sigma_{\underline{0}}'\right),\\
\left(\Sigma_{\underline{y}^{L}}\right)_{jj}&=&\sigma_{y_j^L}^2=\mu_{y_j^L}-\mu_{y_j^L}^2-\sum_{1\leq i\leq N, i\neq j}E\left\{\frac{\partial \,s\left(\underline{z}^L, j\right)}{\partial \left(z_j^L-z_i^L\right)}\right\},\\
\left(\Sigma_{\underline{y}^{L}}\right)_{ij}&=&\sigma_{y_i^Ly_j^L}=-\mu_{y_i^L}\mu_{y_j^L}+E\left\{\frac{\partial \,s\left(\underline{z}^L, j\right)}{\partial \left(z_j^L-z_i^L\right)}\right\}, \quad i\neq j,\\
\left(\Sigma_{\underline{z}^{L}\underline{y}^{L}}\right)_{jj}&=&\sigma_{z_j^Ly_j^L}=\sigma_{z_j^L}^2\sum_{1\leq i\leq N, i\neq j} E\left\{\frac{\partial \,s\left(\underline{z}^L, j\right)}{\partial \left(z_j^L-z_i^L\right)}\right\},\\
\left(\Sigma_{\underline{z}^{L}\underline{y}^{L}}\right)_{ij}&=&\sigma_{z_i^Ly_j^L}=-\sigma_{z_i^L}^2E\left\{\frac{\partial \,s\left(\underline{z}^L, j\right)}{\partial \left(z_j^L-z_i^L\right)}\right\}, \quad i\neq j,
\end{eqnarray*}
where
$$
E\left\{\frac{\partial \,s\left(\underline{z}^L, j\right)}{\partial \left(z_j^L-z_i^L\right)}\right\}\approx
\frac{1}{\sqrt{1+\lambda^2\sigma_{z_j^L}^2+\lambda^2\sigma_{z_i^L}^2}}
\frac{\partial\Phi\left(\frac{\lambda\left(\mu_{z_j^L}-\mu_{z_{\tau}^L}\right)}{\sqrt{1+\lambda^2\sigma_{z_j^L}^2+\lambda^2\sigma_{z_{\tau}^L}^2}}, 1\leq \tau\leq N, \tau\neq j; \mu_{\underline{0}}, \Sigma_{\underline{0}}'\right)}{\partial\left(z_j^L-z_i^L\right)}.
$$
\end{enumerate}
\end{Proposition}
\bigskip

\section{Bayesian Algorithm: Backward Pass}
\medskip

Let $1\leq \ell \leq L$ and $D_{k}^{\ell}=\left\{\left(\underline{y}^{\ell-1}_1, \underline{y}_{1}^{\ell}\right), \ldots, \left(\underline{y}^{\ell-1}_{k}, \underline{y}^{\ell}_{k}\right)\right\}$ be the first $k$ training instances in the layer $\ell$, where $1\leq k\leq m$. For $1\leq j\leq n^{\ell}$, let $p_{k-1}\left(\underline{w}^{\ell}_{j}\right)=p\left(\underline{w}_{j}^{\ell}\,|\,D_{k-1}^{\ell}\right)\approx N\left(\underline{w}^{\ell}_{j}; \mu^{\underline{w}_{j}^{\ell}}_{k-1}, \Sigma^{\underline{w}_{j}^{\ell}}_{k-1}\right)$ be the prior distribution resulting from processing all $k-1$ training instances $D_{k-1}^{\ell}=\left\{\left(\underline{y}^{\ell-1}_1, \underline{y}_{1}^{\ell}\right), \ldots, \left(\underline{y}^{\ell-1}_{k-1}, \underline{y}^{\ell}_{k-1}\right)\right\}$.
Since the posterior distribution cannot be calculated accurately in general, a Gaussian distribution is used for approximating the true posterior, which captures the posterior mean and covariance. Hence updating weights $\underline{w}^{\ell}_{j}$ by means of the so far unseen $k$-th training instance $\left(\underline{y}^{\ell-1}_k, \underline{y}^{\ell}_{k}\right)$ corresponds to calculating the posterior distribution $p_k\left(\underline{w}_{j}^{\ell}\right)=p\left(\underline{w}_{j}^{\ell}\,|\,D_{k}^{\ell}\right)\approx N\left(\underline{w}^{\ell}_{j}; \mu^{\underline{w}_{j}^{\ell}}_{k}, \Sigma^{\underline{w}_{j}^{\ell}}_{k}\right)$.
\medskip

First observe, $p\left(\underline{z}^{\ell} \,|\, \underline{y}_k^{\ell-1}, D_{k-1}^{\ell}\right)\approx N\left(\underline{z}^{\ell}; \mu_{\underline{z}^{\ell}}, \Sigma_{\underline{z}^{\ell}}\right)$ is approximately Gaussian with
$\left(\mu_{\underline{z}^{\ell}}\right)_j=\mu_{z_{j}^{\ell}}=\underline{y}_k^{\ell-1}\mu^{\underline{w}_{j}^{\ell}}_{k-1}$ and
$\left(\Sigma_{\underline{z}^{\ell}}\right)_{jj}=\sigma_{z_{j}^{\ell}}^2
=\underline{y}_k^{\ell-1}\Sigma^{\underline{w}_{j}^{\ell}}_{k-1}\left(\underline{y}_k^{\ell-1}\right)^T$, and $\left(\Sigma_{\underline{z}^{\ell}}\right)_{ij}=0$ for $1\leq i, j\leq n^{\ell}$ and $i\neq j$, where the mean vector $\mu^{\underline{w}_{j}^{\ell}}_{k-1}$ and covariance matrix $\Sigma^{\underline{w}_{j}^{\ell}}_{k-1}$ are given by the prior distribution.
\medskip

By Bayes' theorem,
$$
p\left(\underline{z}^{\ell} \,|\, D_k^{\ell}\right)=\frac{p\left(\underline{y}^{\ell} \,|\, \underline{z}^{\ell}\right)
p\left(\underline{z}^{\ell} \,|\, \underline{y}_{k}^{\ell-1}, D_{k-1}^{\ell}\right)}
{\int\cdots\int p\left(\underline{y}^{\ell} \,|\, \underline{z}^{\ell}\right)p\left(\underline{z}^{\ell} \,|\, \underline{y}_k^{\ell-1}, D_{k-1}^{\ell}\right)dz_1^{\ell}\cdots dz_{n^{\ell}}^{\ell}}.
$$
Assuming that $\underline{y}^{\ell}$ and $\underline{z}^{\ell}$ are jointly Gaussian yields an updated Gaussian approximation, i.e.,
$p\left(\underline{z}^{\ell} \,|\, D_k^{\ell}\right)\approx N\left(\underline{z}^{\ell}; \widetilde{\mu}_{\underline{z}^{\ell}}, \widetilde{\Sigma}_{\underline{z}^{\ell}}\right)$
with the mean vector and variance matrix
\begin{equation}\label{z}
\widetilde{\mu}_{\underline{z}^{\ell}}=\mu_{\underline{z}^{\ell}}+\Sigma_{\underline{z}^{\ell}\underline{y}^{\ell}}
\Sigma_{\underline{y}^{\ell}}^{-1}\left(\underline{y}_{k}^{\ell}-\mu_{\underline{y}^{\ell}}\right)^T \,\,\, {\rm and} \,\,\,
\widetilde{\Sigma}_{\underline{z}^{\ell}}=
\Sigma_{\underline{z}^{\ell}}-\Sigma_{\underline{z}^{\ell}\underline{y}^{\ell}}
\Sigma_{\underline{y}^{\ell}}^{-1}\Sigma_{\underline{z}^{\ell}\underline{y}^{\ell}}^T.
\end{equation}

Now for $1\leq \ell\leq n^{\ell}$ consider the posterior weights
$$
p_k\left(\underline{w}_{j}^{\ell}\right)=\int_{\mathbb{R}} p\left(\underline{w}_{j}^{\ell}, z_{j}^{\ell} \,|\, D_k^{\ell}\right) dz_{j}^{\ell}=\int_{\mathbb{R}} p\left(\underline{w}_{j}^{\ell} \,|\, z_{j}^{\ell}, D_k^{\ell}\right)p\left(z_{j}^{\ell} \,|\, D_k^{\ell}\right)dz_{j}^{\ell}.
$$
Observe $\underline{w}_{j}^{\ell}$ and $z_{j}^{\ell}$ are jointly Gaussian due to the linear mapping. We have
$$
p\left(\underline{w}_{j}^{\ell} \,|\, z_{j}^{\ell}, D_k^{\ell}\right)=N\left(\underline{w}_{j}^{\ell}; \, \mu^{\underline{w}_{j}^{\ell}}_{k-1}+\sigma_{\underline{w}_j^{\ell}z_{j}^{\ell}}
\sigma_{z_j^{\ell}}^{-2}\left(z_j^{\ell}-\mu_{z_j^{\ell}}\right), \, \Sigma^{\underline{w}_j^{\ell}}_{k-1}
-\sigma_{\underline{w}_j^{\ell}z_{i}^{\ell}}\sigma_{z_j^{\ell}}^{-2}\left(\sigma_{\underline{w}_j^{\ell}z_{j}^{\ell}}\right)^T\right).
$$
with the covariance
$$
\sigma_{\underline{w}_j^{\ell}z_{j}^{\ell}}=E\left\{\left(\underline{w}_j^{\ell}-\mu_{k-1}^{\underline{w}_j^{\ell}}\right)
\left(z_j^{\ell}-\mu_{z_j^{\ell}}\right)\right\}
=E\left\{\left(\underline{w}_j^{\ell}-\mu_{k-1}^{\underline{w}_j^{\ell}}\right)
\left(\underline{w}_j^{\ell}-\mu_{k-1}^{\underline{w}_j^{\ell}}\right)^T\right\}\left(\underline{y}_k^{\ell-1}\right)^T
=\Sigma^{\underline{w}_{j}^{\ell}}_{k-1}\left(\underline{y}_k^{\ell-1}\right)^T.
$$
Therefore \, $p_k\left(\underline{w}_{j}^{\ell}\right)=p\left(\underline{w}_{j}^{\ell}\,|\,D_{k}^{\ell}\right)\approx N\left(\underline{w}^{\ell}_{j}; \mu^{\underline{w}_{j}^{\ell}}_{k}, \Sigma^{\underline{w}_{j}^{\ell}}_{k}\right)$, where
\begin{equation}\label{w}
\mu^{\underline{w}_{j}^{\ell}}_{k}=\mu^{\underline{w}_{j}^{\ell}}_{k-1}+\sigma_{\underline{w}_j^{\ell}z_{j}^{\ell}}\sigma_{z_j^{\ell}}^{-2}
\left(\widetilde{\mu}_{z_j^{\ell}}-\mu_{z_j^{\ell}}\right) \quad {\rm and} \quad  \Sigma^{\underline{w}_{j}^{\ell}}_{k}=\Sigma^{\underline{w}_j^{\ell}}_{k-1}
+\sigma_{\underline{w}_j^{\ell}z_{j}^{\ell}}\sigma_{z_j^{\ell}}^{-4}
\left(\widetilde{\sigma}_{z_j^{\ell}}^2-\sigma_{z_j^{\ell}}^2\right)\left(\sigma_{\underline{w}_j^{\ell}z_{j}^{\ell}}\right)^T
\end{equation}
\smallskip

In summary, we have the following algorithms:
\medskip

{\bf Algorithm 1} \, Forward Pass for test input $\underline{y}^0=\underline{x}$
\medskip

\begin{itemize}
\item[1:] for $\ell=1$ to $L$
\item[2:] \qquad Compute $\mu_{\underline{z}^{\ell}}$ and $\Sigma_{\underline{z}^{\ell}}$ by Equation (\ref{zz})
\item[3:] \qquad if $\ell=1$ to $L-1$
\item[4:] \qquad\qquad {\it pwl:} \, Calculate mean vector $\mu_{\underline{y}^{\ell}}$, $\Sigma_{\underline{y}^{\ell}}$, and $\Sigma_{\underline{z}^{\ell}\underline{y}^{\ell}}$ using Proposition 3.1 (1)
\item[5:] \qquad\qquad replace $\underline{y}^{\ell}$ by $\mu_{\underline{y}^{\ell}}$
\item[6:] \qquad if $\ell=L$
\item[7:] \qquad\qquad {\it softmax:} \, Calculate mean vector $\mu_{\underline{y}^{L}}$, $\Sigma_{\underline{y}^{L}}$, and $\Sigma_{\underline{z}^{L}\underline{y}^{L}}$ using Proposition 3.1 (2)
\item[8:] end for
\item[9:] Return $\left(\mu_{\underline{z}^{\ell}}, \Sigma_{\underline{z}^{\ell}}, \mu_{\underline{y}^{\ell}}, \Sigma_{\underline{y}^{\ell}}, \Sigma_{\underline{z}^{\ell}\underline{y}^{\ell}}, \underline{y}^{\ell}\right)$ for $1\leq \ell\leq L$
\end{itemize}
\medskip

{\bf Algorithm 2} \, Backward Pass for updating weights with data $D=\left\{\underline{x}_k, \underline{y}_k\right\}_{k=1}^m$
\medskip

\begin{itemize}
\item[1:] for $\ell=1$ to $L$ and $1\leq j\leq n^{\ell}$
\item[2:] \qquad Initialize weight distribution with mean vector $\mu_0^{\underline{w}_j^{\ell}}$ and covariance matrix $\Sigma_0^{\underline{w}_j^{\ell}}$
\item[3:] end for
\item[4:] for each training instance $\left(\underline{x}_k, y_k\right) \in D$ do
\item[5:] \qquad Perform Algorithm 1 to obtain $\left(\mu_{\underline{z}^{\ell}}, \Sigma_{\underline{z}^{\ell}}, \mu_{\underline{y}^{\ell}}, \Sigma_{\underline{y}^{\ell}}, \Sigma_{\underline{z}^{\ell}\underline{y}^{\ell}}, \underline{y}^{\ell}\right)$ for $1\leq \ell\leq L$
\item[6:] \qquad For $\ell=L, L-1, \ldots, 1$
\item[7:] \qquad Calculate $\widetilde{\mu}_{\underline{z}^{\ell}}$ and $\widetilde{\Sigma}_{\underline{z}^{\ell}}$ using Equation (\ref{z})
\item[8:] \qquad Update the mean vector $\mu_k^{\underline{w}_j^{\ell}}$ and covariance matrix $\Sigma_k^{\underline{w}_j^{\ell}}$ for $1\leq j\leq n^{\ell}$ using Equation (\ref{w})
\item[9:] end for
\item[10:] Return $\left(\mu_m^{\underline{w}_j^{\ell}}, \Sigma_m^{\underline{w}_j^{\ell}}\right)$ for $1\leq j\leq n^{\ell}$ and $1\leq \ell\leq L$
\end{itemize}

\section{Experiment results}
\medskip

Let $D=\{(\underline{x}_1, \underline{y}_1), \ldots, (\underline{x}_{25}, \underline{y}_{25})\}$ be a training data set, where $\underline{x}_k=[x_{k1} \,\, x_{k2}]$ for $1\leq k\leq 25$ are generated uniformly at random over the two-dimensional area $[-2, 2]\times [-2, 2]$. The data points $\underline{x}=[x_1\,\,x_2]\in \mathbb{R}^2$ are assigned to one of the three classes according to
\begin{equation}\label{f}
\underline{y}=\begin{cases} [1,0],\qquad {\rm if} \quad (x_1+x_2)>0\,\,\&\,\,(-x_1+x_2)>0,\\
[0, 1],\qquad {\rm if} \quad (x_1+x_2)<0\,\,\&\,\,(-x_1+x_2)<0,\\
[0,0], \qquad {\rm otherwise}. \end{cases}
\end{equation}

We are going to apply the proposed Bayesian algorithm to train the following single layer neural network for 3-class classification:
\newpage
$$\xymatrix@R=0.3cm{
  x_1, x_2 \ar[r]^{$\underline{w}_1$} &  \fbox{$z_1$}\ar[r]^{$s$}\ar[dr] & \fbox{$y_1$} \\
  x_1, x_2 \ar[r]^{$\underline{w}_2$} &  \fbox{$z_2$}\ar[r]^{$s$}\ar[ur] & \fbox{$y_2$} \\
  x_1, x_2 \ar[r]^{$\underline{w}_3$} &  \fbox{$z_3$}\ar[ur]\ar[uur] &   }
$$

The weight distributions are initialized with $\underline{w}_1\sim
N\left(\underline{\mu}_0^{\underline{w}_1}, \Sigma_0^{\underline{w}_1}\right)$,
where $\underline{\mu}_0^{\underline{w}_1}=\left[1\,\,0\right]^T$ and $\Sigma_0^{\underline{w}_1}=I_2$,
$\underline{w}_2\sim
N\left(\underline{\mu}_0^{\underline{w}_2}, \Sigma_0^{\underline{w}_2}\right)$,
where $\underline{\mu}_0^{\underline{w}_2}=\left[0\,\,1\right]^T$ and $\Sigma_0^{\underline{w}_2}=I_2$,
and $\underline{w}_3\sim
N\left(\underline{\mu}_0^{\underline{w}_3}, \Sigma_0^{\underline{w}_3}\right)$,
where $\underline{\mu}_0^{\underline{w}_2}=\left[1\,\,1\right]^T$ and
$\Sigma_0^{\underline{w}_2}=I_2$. Here $I_2$ is the $2\times 2$ identity matrix.

We display the evolutions of the predictive means and the variances $\mu_{y_1}$, $\mu_{y_2}$, $\sigma_{y_1}^2$ and $\sigma_{y_2}^2$ in the following graphs. Figures (1)-(3) and Figures (4)-(6) respectively show the evolutions of the predictive means $\mu_{y_1}$ and  $\mu_{y_2}$. It can been seen that the initially rather high indifference becomes continuously sharper and the
algorithm is able to correctly learn the decision boundary being defined by Equation (\ref{f}) for both classes one and two. Figures (7-9) and Figures (10-12) exhibit the evolutions of variances $\sigma_{y_1}^2$ and $\sigma_{y_2}^2$ separately. These figures provide
uncertainty quantification of the Bayesian algorithm.

\hspace{-0.2in}
\begin{minipage}{0.33\linewidth}
\includegraphics[width = \linewidth]{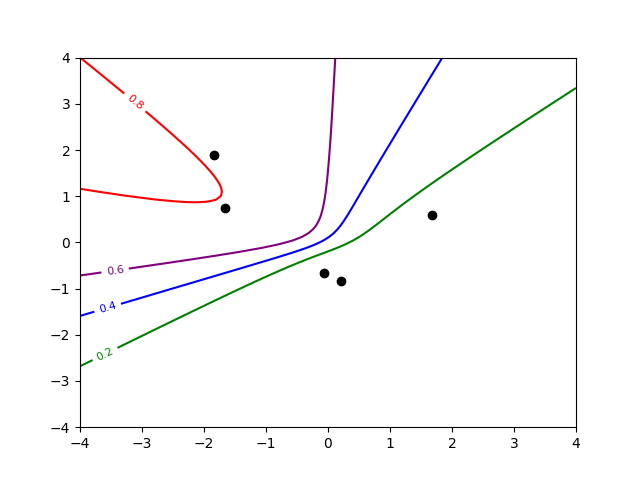}

\hspace{1in}(1)
\end{minipage}
\begin{minipage}{0.33\linewidth}
\includegraphics[width = \linewidth]{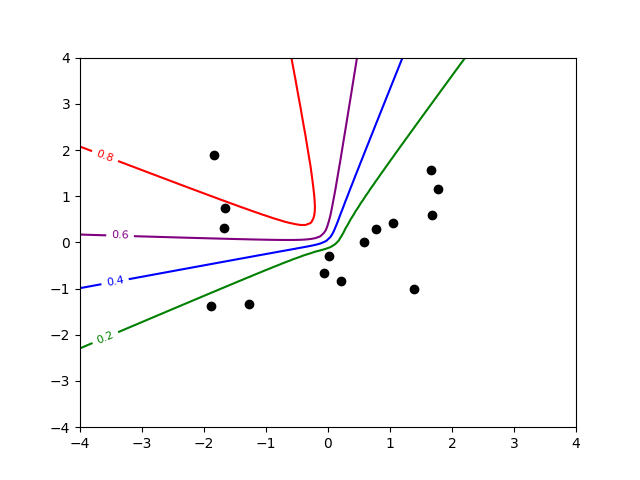}

\hspace{1in}(2)
\end{minipage}
\begin{minipage}{0.33\linewidth}
\includegraphics[width = \linewidth]{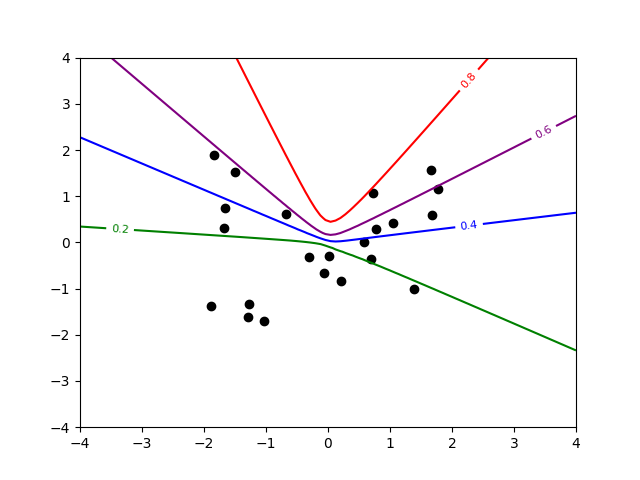}

\hspace{1in}(3)
\end{minipage}

\hspace{-0.2in}
\begin{minipage}{0.33\linewidth}
\includegraphics[width = \linewidth]{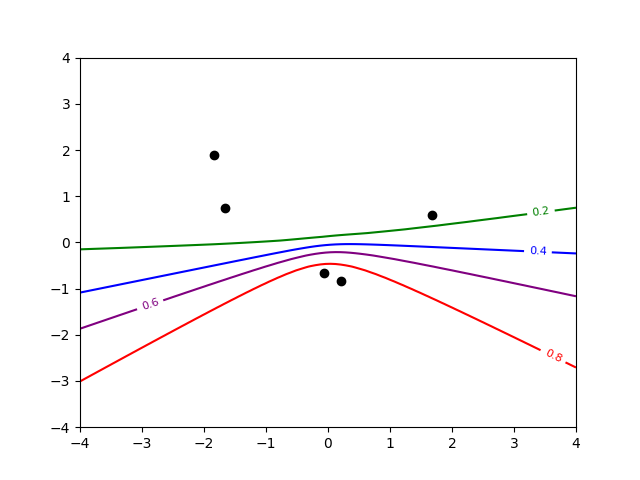}

\hspace{1in}(4)
\end{minipage}
\begin{minipage}{0.33\linewidth}
\includegraphics[width = \linewidth]{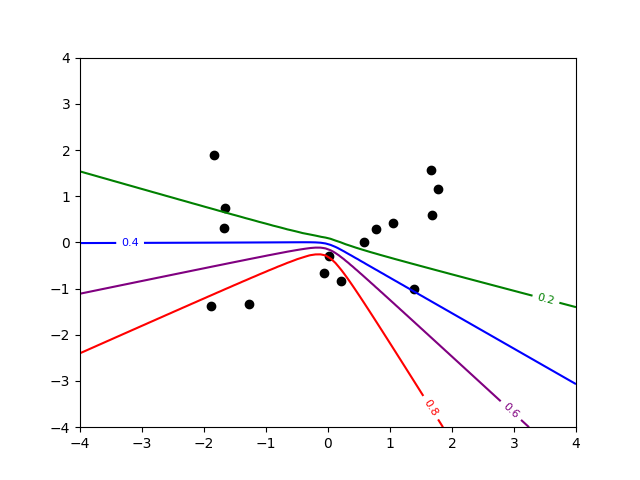}

\hspace{1in}(5)
\end{minipage}
\begin{minipage}{0.33\linewidth}
\includegraphics[width = \linewidth]{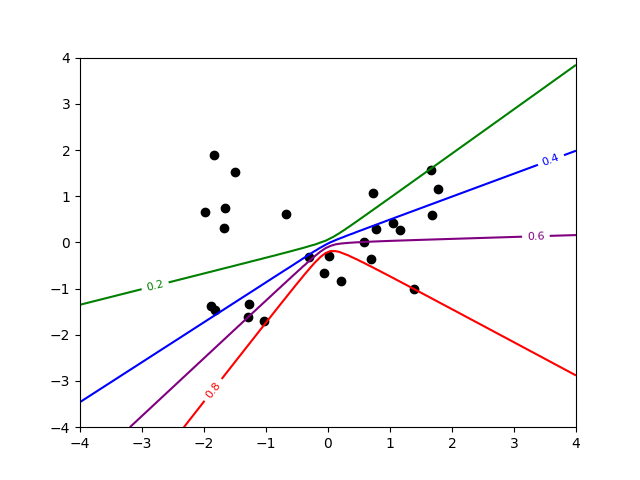}

\hspace{1in}(6)
\end{minipage}

\hspace{-0.2in}
\begin{minipage}{0.33\linewidth}
\includegraphics[width = \linewidth]{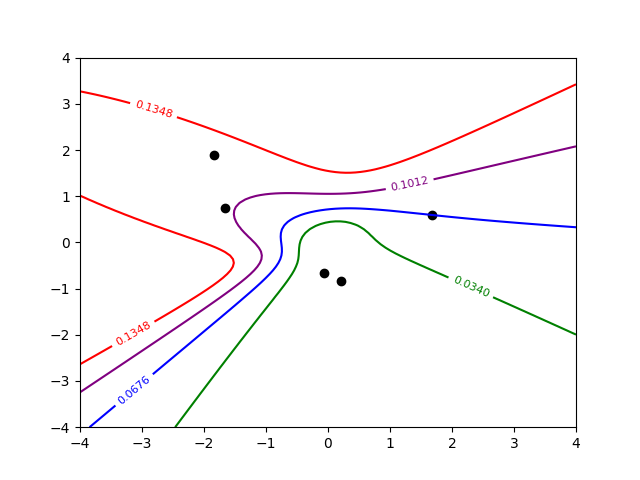}

\hspace{1in}(7)
\end{minipage}
\begin{minipage}{0.33\linewidth}
\includegraphics[width = \linewidth]{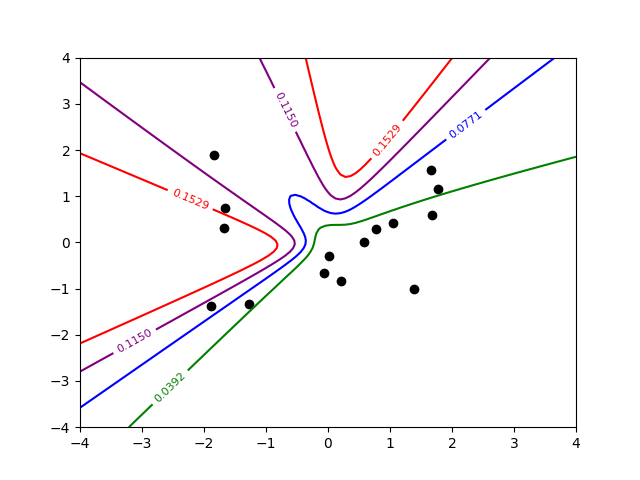}

\hspace{1in}(8)
\end{minipage}
\begin{minipage}{0.33\linewidth}
\includegraphics[width = \linewidth]{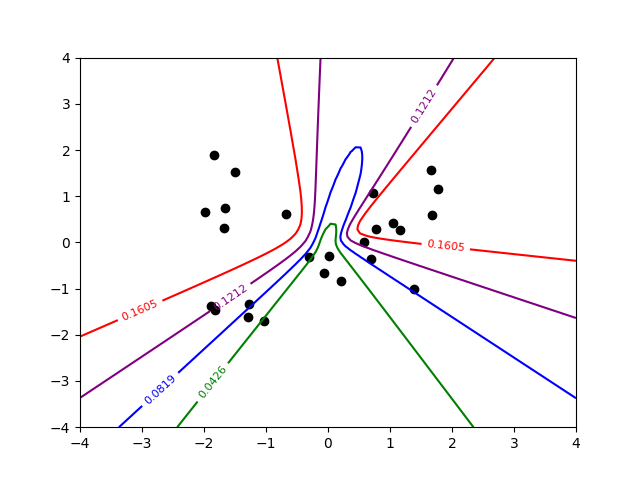}

\hspace{1in}(9)
\end{minipage}

\hspace{-0.2in}
\begin{minipage}{0.33\linewidth}
\includegraphics[width = \linewidth]{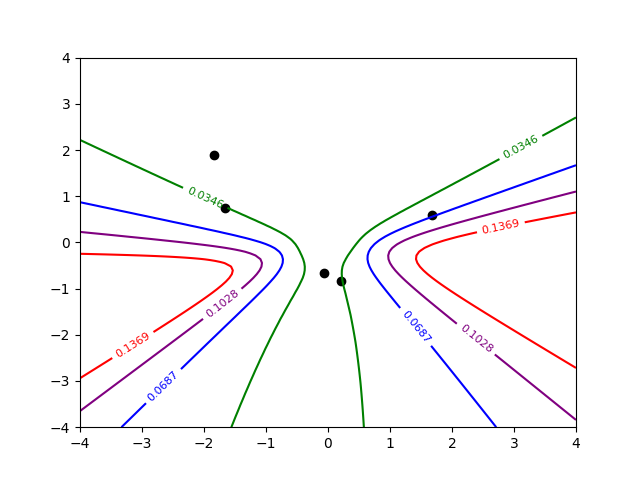}

\hspace{1in}(10)
\end{minipage}
\begin{minipage}{0.33\linewidth}
\includegraphics[width = \linewidth]{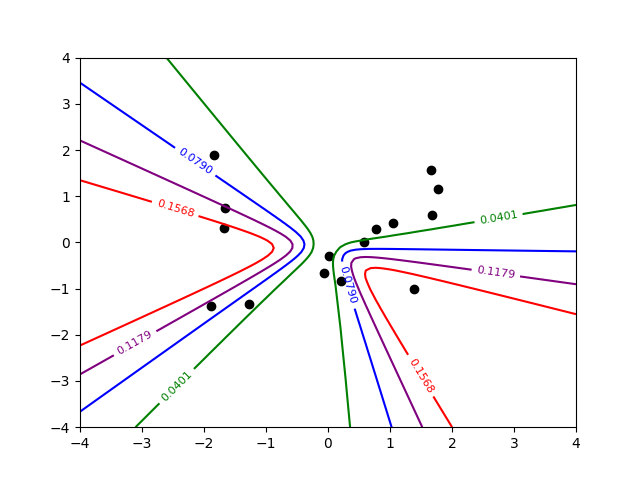}

\hspace{1in}(11)
\end{minipage}
\begin{minipage}{0.33\linewidth}
\includegraphics[width = \linewidth]{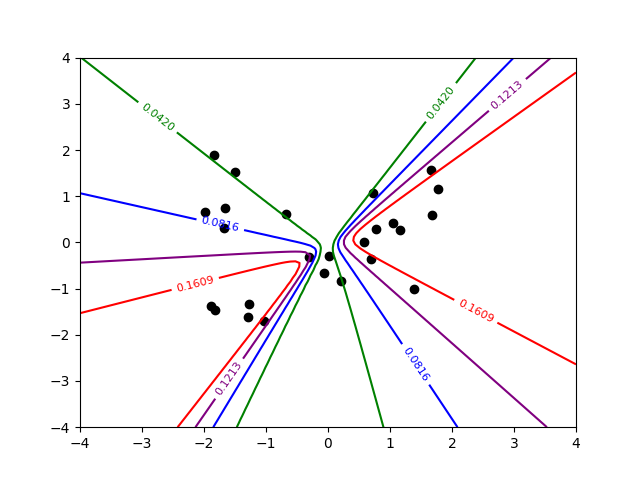}

\hspace{1in}(12)
\end{minipage}
\bigskip

\section{Conclusions}
\medskip

In this paper, we introduce a new type of Bayesian learning algorithm for fully connected neural networks for multi-class classification. This algorithm trains the weights by means of Bayesian inference in closed form without the need of gradient calculation and Monte Carlo sampling. It also provides uncertainty of the prediction by calculating the probability distribution of the output. The learning can be performed sequentially which allows for on-line learning and realtime applications. This idea is motivated by the work of Marco Huber in \cite{H} who proposed Bayesian perceptron for a single layer network for binary classification.  In the future, we would like to investigate whether this algorithm can be generalized to other deep learning neural networks. We also plan to do more experiments to test this algorithm and compare it with traditional Bayesian learning algorithms such as variational inference, dropout, and Kalman filters.

\section{Appendix}
\medskip

Let $\underline{z}=[z_1 \,\, \ldots \,\, z_n]\sim N\left(\mu_{\underline{z}}, \Sigma_{\underline{z}}\right)$ be $n$ Gaussian random variables with mean vector $\mu_{\underline{z}}$ and covariance matrix $\Sigma_{\underline{z}}$, where $\left(\mu_{\underline{z}}\right)_j=\mu_{z_j}$ and
$\left(\Sigma_{\underline{z}}\right)_{jj}=\sigma_{z_j}^2, \left(\Sigma_{\underline{z}}\right)_{ij}=0$ for $1\leq i, j\leq n$ and $i\neq j$.
Fix $j$ such that $1\leq j\leq n$.
Let $\underline{u}=[u_{\tau} \,|\, 1\leq \tau\leq n, \tau\neq j]\sim N\left(\mu_{\underline{0}}, \Sigma_{\underline{0}}\right)$ (see Equations (\ref{mean}) and (\ref{variance}) in Section 2 for $\mu_{\underline{0}}$ and $\Sigma_{\underline{0}}$).
Then $\underline{v}=[u_{\tau}-z_j+z_{\tau} \,|\, 1\leq \tau\leq n, \tau\neq j]\sim N\left(\mu_{\underline{v}}, \Sigma_{\underline{v}}\right)$, where
$$
\mu_{\underline{v}}=\left[\begin{array}{c}-\mu_{z_j}+\mu_{z_{\tau}}\,|\, 1\leq \tau\leq n, {\tau}\neq j\end{array}\right],
\left(\Sigma_{\underline{v}}\right)_{\tau\tau}=1+\sigma_{z_j}^2+\sigma_{z_{\tau}}^2, 1\leq \tau\leq n, \tau\neq j,
$$
and
$$
\left(\Sigma_{\underline{v}}\right)_{\tau \tau'}=\begin{cases}\rho_{\tau\tau'}+\sigma_{z_j}^2, \quad 1\leq \tau, \tau'\leq j-1, \tau\neq \tau',\\
\rho_{\tau\left(\tau'-1\right)}+\sigma_{z_j}^2, \quad 1\leq \tau\leq j-1, j+1\leq \tau'\leq n,\\
\rho_{\left(\tau-1\right)\tau'}+\sigma_{z_j}^2, \quad j+1\leq \tau\leq n, 1\leq \tau'\leq j-1,\\
\rho_{\left(\tau-1\right)\left(\tau'-1\right)}+\sigma_{z_j}^2, \quad j+1\leq \tau, \tau'\leq n, \tau\neq \tau'.
\end{cases}
$$
\smallskip

Let $\underline{w}=\left[\frac{v_{\tau}-\mu_{v_{\tau}}}{\sigma_{v_{\tau}}} \,|\, 1\leq \tau\leq n, \tau\neq j\right]$ be the standardized version of $\underline{v}$. Then $\underline{w}\sim N\left(\mu_{\underline{0}}, \Sigma_{\underline{0}}'\right)$, where
\begin{equation}
\mu_{\underline{0}}=\left[\begin{array}{ccc}0&\cdots&0\end{array}\right]_{1\times (n-1)}, \,\,
\left(\Sigma_{\underline{0}}'\right)_{\tau\tau}=1, \, 1\leq \tau\leq n, \tau\neq j,
\end{equation}
and
\begin{equation}
\left(\Sigma_{\underline{0}}'\right)_{\tau\tau'}=\begin{cases}\frac{\rho_{\tau\tau'}+\sigma_{z_j}^2}
{\sqrt{\left(1+\sigma_{z_j}^2+\sigma_{z_{\tau}}^2\right)
\left(1+\sigma_{z_{j}}^2+\sigma_{z_{\tau'}}^2\right)}}, \quad 1\leq \tau, \tau'\leq j-1, \tau\neq \tau',\\
\frac{\rho_{\tau\left(\tau'-1\right)}+\sigma_{z_j}^2}{\sqrt{\left(1+\sigma_{z_j}^2+\sigma_{z_{\tau}}^2\right)
\left(1+\sigma_{z_j}^2+\sigma_{z_{\tau'}}^2\right)}}, \quad 1\leq \tau\leq j-1, j+1\leq \tau'\leq n,\\
\frac{\rho_{\left(\tau-1\right)\tau'}+\sigma_{z_j}^2}{\sqrt{\left(1+\sigma_{z_j}^2+\sigma_{z_{\tau}}^2\right)
\left(1+\sigma_{z_j}^2+\sigma_{z_{\tau'}}^2\right)}}, \quad j+1\leq \tau\leq n, 1\leq \tau'\leq j-1,\\
\frac{\rho_{\left(\tau-1\right)\left(\tau'-1\right)}+\sigma_{z_j}^2}{\sqrt{\left(1+\sigma_{z_j}^2+\sigma_{z_{\tau}}^2\right)
\left(1+\sigma_{z_j}^2+\sigma_{z_{\tau'}}^2\right)}}, \quad j+1\leq \tau, \tau'\leq n, \tau\neq \tau'.
\end{cases}
\end{equation}
\smallskip

By Equation (\ref{DefinitionPhi}) in Section 2,
$$\Phi\left(z_j-z_{\tau}, 1\leq \tau\leq n, \tau\neq j; \mu_{\underline{0}}, \Sigma_{\underline{0}}\right)
$$
$$
=\int_{-\infty}^{z_j-z_n}\cdots\int_{-\infty}^{z_j-z_1}
N\left(t_{\tau}, 1\leq \tau\leq n, \tau\neq j; \mu_{\underline{0}}, \Sigma_{\underline{0}}\right)dt_1\cdots \widehat{dt_j}\cdots dt_{n}
$$
where $\widehat{dt_j}$ means $dt_j$ is skipped.
\medskip

\begin{Proposition} \, With $\mu_{\underline{0}}, \Sigma_{\underline{0}}$, and $\Sigma_{\underline{0}}'$ defined as above, one has the following equations:

\begin{enumerate} \label{phi}
\item $$\int_{\mathbb{R}}\cdots\int_{\mathbb{R}}\Phi\left(z_j-z_{\tau}, 1\leq \tau\leq n, \tau\neq j; \mu_{\underline{0}}, \Sigma_{\underline{0}}\right)N\left(\underline{z}; \mu_{\underline{z}}, \Sigma_{\underline{z}}\right)dz_1\cdots dz_n
$$
$$
=\Phi\left(\frac{\mu_{z_j}-\mu_{z_{\tau}}}{\sqrt{1+\sigma_{z_j}^2+\sigma_{z_{\tau}}^2}}, 1\leq \tau\leq n, \tau\neq j; \mu_{\underline{0}}, \Sigma_{\underline{0}}'\right).$$
\item $$\int_{\mathbb{R}}\cdots\int_{\mathbb{R}}\frac{\partial \,\Phi\left(z_j-z_{\tau}, 1\leq \tau\leq n, \tau\neq j; \mu_{\underline{0}}, \Sigma_{\underline{0}}\right)}{\partial (z_j-z_i)}N\left(\underline{z}; \mu_{\underline{z}}, \Sigma_{\underline{z}}\right)dz_1\cdots dz_n
$$
$$
=\frac{1}{\sqrt{1+\sigma_{z_j}^2+\sigma_{z_i}^2}}
\frac{d\Phi}{d(z_j-z_i)}\left(\frac{\mu_{z_j}-\mu_{z_{\tau}}}{\sqrt{1+\sigma_{z_j}^2+\sigma_{z_{\tau}}^2}}, 1\leq \tau\leq n, \tau\neq j; \mu_{\underline{0}}, \Sigma_{\underline{0}}'\right).
$$
\end{enumerate}
\end{Proposition}

\noindent
{\bf Proof.} \,
Since
$$
\Phi\left(t_j-t_{\tau}, 1\leq \tau\leq n, \tau\neq j; \mu_{\underline{0}}, \Sigma_{\underline{0}}\right)=
P\left(u_{\tau}\leq z_j-z_{\tau}, 1\leq \tau\leq n, \tau\neq j \,|\, \underline{z}=[t_1 \,\, \ldots \,\, t_n]\right),
$$
one has
\begin{eqnarray}
&&\int_{\mathbb{R}}\cdots\int_{\mathbb{R}}\Phi\left(z_j-z_{\tau}, 1\leq \tau\leq n, \tau\neq j; \mu_{\underline{0}}, \Sigma_{\underline{0}}\right)N\left(\underline{z}; \mu_{\underline{z}}, \Sigma_{\underline{z}}\right)dz_1\cdots dz_n \nonumber \\
&=&\int_{\mathbb{R}}\cdots\int_{\mathbb{R}}\Phi\left(t_j-t_{\tau}, 1\leq \tau\leq n, \tau\neq j; \mu_{\underline{0}}, \Sigma_{\underline{0}}\right)N\left(\underline{t}; \mu_{\underline{z}}, \Sigma_{\underline{z}}\right)dt_1\cdots dt_n \nonumber \\
&=&P\left(u_{\tau}\leq z_j-z_{\tau}, 1\leq \tau\leq n, \tau\neq j\right)=P\left(v_{\tau}\leq 0, 1\leq \tau\leq n, \tau\neq j\right) \nonumber
\end{eqnarray}
\begin{eqnarray}
&=&P\left(w_{\tau}\leq \frac{\mu_{z_j}-\mu_{z_{\tau}}}{\sqrt{1+\sigma_{z_j}^2+\sigma_{z_{\tau}}^2}}, 1\leq \tau\leq n, \tau\neq j\right) \nonumber \\
&=&\Phi\left(\frac{\mu_{z_j}-\mu_{z_{\tau}}}{\sqrt{1+\sigma_{z_j}^2+\sigma_{z_{\tau}}^2}}, 1\leq \tau\leq n, \tau\neq j; \mu_{\underline{0}}, \Sigma_{\underline{0}}'\right). \nonumber
\end{eqnarray}
Moreover, since
\begin{eqnarray*}
&&\frac{\partial \,\Phi\left(t_j-t_{\tau}, 1\leq \tau\leq n, \tau\neq j; \mu_{\underline{0}}, \Sigma_{\underline{0}}\right)}{\partial (z_j-z_i)}\\
&=&P\left(u_{i}=t_j-t_{i}, u_{\tau}\leq t_j-t_{\tau}, 1\leq \tau\leq n, \tau\neq j, i \,|\, \underline{z}=[t_1 \,\, \ldots \,\, t_n]\right),
\end{eqnarray*}
we have
\begin{eqnarray}
&&\int_{\mathbb{R}}\cdots\int_{\mathbb{R}}\frac{\partial \,\Phi\left(z_j-z_{\tau}, 1\leq \tau\leq n, \tau\neq j; \mu_{\underline{0}}, \Sigma_{\underline{0}}\right)}{\partial (z_j-z_i)}N\left(\underline{z}; \mu_{\underline{z}}, \Sigma_{\underline{z}}\right)dz_1\cdots dz_n
\nonumber \\
&=&\int_{\mathbb{R}}\cdots\int_{\mathbb{R}}\frac{\partial \,\Phi\left(t_j-t_{\tau}, 1\leq \tau\leq n, \tau\neq j; \mu_{\underline{0}}, \Sigma_{\underline{0}}\right)}{\partial (z_j-z_i)}N\left(\underline{t}; \mu_{\underline{z}}, \Sigma_{\underline{z}}\right)dt_1\cdots dt_n
\nonumber \\
&=&P\left(u_i=z_j-z_i, u_{\tau}\leq z_j-z_{\tau}, 1\leq \tau\leq n, \tau\neq j, i\right) \nonumber \\
&=&P\left(v_i=0, v_{\tau}\leq 0, 1\leq \tau\leq n, \tau\neq j, i\right) \nonumber \\
&=&P\left(w_{i}= \frac{\mu_{z_j}-\mu_{z_{i}}}{\sqrt{1+\sigma_{z_j}^2+\sigma_{z_{i}^2}}}, w_{\tau}\leq \frac{\mu_{z_j}-\mu_{z_{\tau}}}{\sqrt{1+\sigma_{z_j}^2+\sigma_{z_{\tau}}^2}}, 1\leq \tau\leq n, \tau\neq j, i\right) \nonumber \\
&=&\frac{1}{\sqrt{1+\sigma_{z_j}^2+\sigma_{z_i}^2}}
\frac{d\Phi}{d(z_j-z_i)}\left(\frac{\mu_{z_j}-\mu_{z_{\tau}}}{\sqrt{1+\sigma_{z_j}^2+\sigma_{z_{\tau}}^2}}, 1\leq \tau\leq n, \tau\neq j; \mu_{\underline{0}}, \Sigma_{\underline{0}}'\right). \nonumber
\end{eqnarray}

\vskip 0.5in
\bibliographystyle{amsalpha}

\end{document}